\relax
\documentclass[letterpaper]{article} 
\usepackage{aaai21}  
\usepackage{times}  
\usepackage{helvet} 
\usepackage{courier}  
\usepackage[hyphens]{url}  
\usepackage{graphicx} 
\usepackage{natbib}  
\usepackage{caption} 
\usepackage[switch]{lineno}

\usepackage{booktabs}
\usepackage[table,xcdraw]{xcolor}

\usepackage{amsmath}
\usepackage{amssymb}

\title{Neuro-symbolic Neurodegenerative Disease Modeling\\as Probabilistic Programmed Deep Kernels}

\author {
    Alexander Lavin \\
}

\affiliations{
    Latent Sciences \\
    lavin@latentsci.com
}

\begin{document}


\maketitle

\begin{abstract}
We present a probabilistic programmed deep kernel learning approach to personalized, predictive modeling of neurodegenerative diseases. Our analysis considers a spectrum of neural and symbolic machine learning approaches, which we assess for predictive performance and important medical AI properties such as interpretability, uncertainty reasoning, data-efficiency, and leveraging domain knowledge. 
Our Bayesian approach combines the flexibility of Gaussian processes with the structural power of neural networks to model biomarker progressions, without needing clinical labels for training.
We run evaluations on the problem of Alzheimer's disease prediction, yielding results that surpass deep learning in both accuracy and timeliness of predicting neurodegeneration, and with the practical advantages of Bayesian non-parametrics and probabilistic programming.

\end{abstract}

\section{Introduction}

Accurate prediction of disease trajectories is critical for informing decisions, particularly in the earliest phases of disease where action can be impactful. Machine learning methods show promise in diagnostics, but mainstream neural network methods struggle to model longitudinal disease biomarkers to predict disease trajectories. Even more, they are ineffective with sparse and limited data that is common in clinical cohorts, and the deep learning methods are difficult to trust due to lack of interpretable mechanisms.
Probabilistic generative models, on the other hand, are well-suited to the task of disease modeling.
This class of methods allows quantification and reasoning with uncertainties, is robust to sparse and noisy data, yields interpretable models and predictions, and can utilize prior information and structure encoded by domain experts.
Probabilistic programming, a relatively new area at the intersection of artificial intelligence and programming languages, provides the ideal representation of these models in software.
A probabilistic model in the form of probabilistic program code describes the data-generating process from unknown latent variables to observed data, and in general the languages enable automatic inference over arbitrary programs.
The types of models that may be written as probabilistic programs include not only basic Bayesian networks and graphical models, but also ones more expressive and flexible, such as non-parametric models and graphical models with dynamic structure.

There have been many successful uses of probabilistic programming and neural networks in medical AI. Yet predictive modeling neurodegenerative diseases presents even more challenges, mainly due to the complex nature and heterogeneity of these conditions. For instance, Alzheimer's disease is characterized by dozens of biomarkers ranging from spinal fluid proteins and genetic precursors to psychophysical assessments and brain MRI, the nature of which vary widely across the population of patients. Even more, there are potentially dozens of subtypes of the disease, and the underlying mechanisms remain relatively unknown. This is compounded by the data regime, where measurements are sparse, cohorts are relatively small, and clinical labels are subjective. To approach the problem of predictive modeling neurodegeneration, we would ideally leverage the representational learning capacity of neural networks, while making use of symbolic methods that operate with little data and enable experts to encode rich domain knowledge.

We propose such a neuro-symbolic approach that integrates the flexibility of Bayesian non-parametrics, the feature learning of neural networks, and the expressive capabilities of probabilistic programming. Our main contributions are the following:

\begin{itemize}
\item We build on the work of deep kernel learning \cite{Wilson2016DeepKL} to develop \textit{probabilistic programmed deep kernel learning (PP-DKL)}, which enables us to improve the biomarker modeling with ``kernel warping" and monotonic Gaussian processes.
\item We discuss and evaluate a spectrum of machine learning approaches towards neurodegenerative disease modeling: from data-driven neural networks to domain-engineered graphical models and statistical methods.
\item We show our PP-DKL method achieves state-of-the-art results on predicting neurodegeneration in Alzheimer's disease, along with practical advantages such as interpretability and uncertainty reasoning. Further, the results imply PP-DKL can predict neurodegeneration earlier than other methods.
\end{itemize}

\section{Background}

\subsection{Probabilistic Programming}
\label{backPP}

\textit{Probabilistic programming} is a paradigm that equates probabilistic generative models with executable programs, 
where an inference backend takes programs and observed data to generate inference results.
Probabilistic programming languages (PPLs) enable one to leverage the power of programming languages to create rich and complex models, and use built–in inference algorithms that can operate on any model written in the language.

A PPL is typically implemented as an extension of an existing Turing complete programming language 
(such as Python \cite{Bingham2019PyroDU}, C/C++ \cite{Paige2014ACT, ProbProg18}, and Julia \cite{Ge2018TuringAL})
with operations for sampling and conditioning of random variables.
This approach yields unrestricted (universal) PPLs~\cite{Gordon2014ProbabilisticP} that can operate on arbitrary programs~\cite{Goodman2008ChurchAL, Pfeffer2009FigaroA, Mansinghka2014VentureAH, Wood-AISTATS-2014, GoodmanStuhlm14}.
Another class of PPL is more restrictive and thus computationally efficient, where constraining the set of expressible models ensures that particular inference algorithms can be efficiently applied~\cite{Lunn2009TheBP, winn2009probabilistic, Milch2005BLOGPM, Tran2016EdwardAL}.

Core to PPL design is the decoupling of modeling and inference; model code is generally concise and modular, and inference algorithms are capable of operating on arbitrary programs. 
These general-purpose ``inference engines'' are typically implemented as forward-simulation sampling methods (namely Sequential Monte Carlo and particle MCMC \cite{Andrieu2010ParticleMC}) or variational inference. 
Inference engines for universal probabilistic programming languages use forward-simulation based sampling methods, which is the basis of our work.

A model program specifies a joint distribution \(p(\mathbf{y},\mathbf{x})\) over data \(\mathbf{y}\) and variables \(\mathbf{x}\). Inference aims to characterize the distribution of \textit{program execution traces}: a map from random choices to their specific values.
In other words, a generative model is compiled to a form that can be interpreted by an inference engine, which then outputs some characterization of the posterior such as a series of samples.
Formally the probability of an execution trace can be defined as
\[ p(\mathbf{y},\mathbf{x}) = \prod_{n = 1}^{N} p(y_{n}|\zeta_{g_{n}},\mathbf{x}_{n})p(\mathbf{x}_{n}|\mathbf{x}_{n-1}) \] 
\noindent
where \(\mathbf{y}_{n}\) is the \(n\)-th observed data point, \(p(y_{n}|\zeta_{g_{n}})\) is its normalized likelihood, \(\zeta_{g_{n}}(\mathbf{x}_{n})\) is the program argument with random procedures \({g_{n}}(\mathbf{x}_{n})\), \(\mathbf{x}_{n}\) is the ordered set of all random choices with \(p(\mathbf{x}_{n}|\mathbf{x}_{n-1})\) its normalized prior probability, and \(\mathbf{x}\) and \(\mathbf{y}\) are the sets of all latent and observing random procedures, respectively \cite{Wood-AISTATS-2014}.

For more detailed derivation and further information on probabilistic programming we refer the reader to the recent tutorial of \citet{Meent2018AnIT}.

\subsection{Neurodegenerative Disease Modeling}

Neurodegenerative disorders such as Alzheimer's disease are characterised by the progressive pathological alteration of the brain's biochemical processes and morphology, leading to irreversible impairment of cognitive functions. These diseases are uniquely challenging to model computationally due to heterogeneous biological pathways \cite{Beckett2015TheAD, Hardy2002TheAH}, complex temporal patterns \cite{Jedynak2015ACM, Donohue2014EstimatingLM}, and diverse interactions \cite{Froelich2017IncrementalVO, Pascoal2017SynergisticIB}.
The driving aim for longitudinal modeling Alzheimer's disease is to provide predictions that enable early detection of at-risk subjects and possibly reveal latent disease mechanisms.

Statistical techniques for longitudinal disease modeling are ubiquitous, most commonly generalized linear or nonlinear mixed-effects models \cite{Laird1982RandomeffectsMF, Lindstrom1990NonlinearME}, and various survival models such as the Cox proportional hazards \cite{cox92}. 
These models often assume linearity and proportionality in the underlying stochastic disease process, which makes them ill-suited to model the complexities of neurodegenerative processes.

More promising statistical approaches apply joint mixed-effects models to uncover longitudinal trends across pathological abnormalities, notably the approach of \citet{Li2019BayesianLT} that constructs a generative model for individual biomarker progressions.
More general is the continuous-time HMM for disease progression from \citet{Liu2015EfficientLO}, specifically to account for irregular sampling of longitudinal data and hidden state-transition times. Their approach yields interpretable trajectories as state transition trends (as does ours), which is a rather useful research and prognostic tool, but the method does not provide predictive modeling.

Bayesian non-parametric methods are promising, notably Gaussian Processes as powerful probabilistic predictors of dementia risk \cite{Peterson2017PersonalizedGP, Ziegler2014IndividualizedGP, Lorenzi2017DiseasePM, Hyun2016STGPSG, Rudovic2019MetaWeightedGP}. For the most part these approaches rely heavily on cognitive assessment scores and clinical status, and mainly model population-level statistics. The result is inability to model early phases of Alzheimer's, nor individual-specific pathologies.

Neural network approaches, as general-purpose function approximators, may be well-suited to the complexities inherent in modeling Alzheimer’s disease. Using deep convolutional neural networks, state-of-the-art vision-based results were shown by \citet{Ding2019ADL}, but with subpar performance in early neurodegeneration, let alone the presymptomatic phases. 
We hypothesize restricting biomarkers to the visual modality sets a bound on how early a model can predict neurodegeneration. Similarly, clinical measures such as cognitive assessments are far too late; research suggests that brain changes associated with the disease begin upwards of 20 years before cognitive symptoms may appear \cite{Villemagne2013AmyloidD}.
Match-Net \cite{Jarrett2020DynamicPI} is a promising deep learning approach that aggregates the biomarker modalities into one deep network, and models longitudinal dependencies with temporal convolutions. Even more, to help address data sparsity issues that are common in neurodegenerative patient populations, the model learns correlations between missing data and disease progression. The authors show state-of-the-art results on the task of Alzheimer's risk scoring, but in a fully supervised way towards an \textit{ill-defined task}. 
That is, most existing approaches focus on modeling patients based on their clinical status, i.e. the clinical diagnosis categorizing them into one of the three main stages of Alzheimer's: cognitively normal (CN), mild cognitive impairment (MCI), and Alzheimer's dementia (AD). These labels are largely subjective and qualitative; misdiagnosis is pervasive in practice \cite{Jobke2018SetbacksIA}. We stress that any machine learning model trained on Alzheimer's clinical labels is thus ill-defined.

Thus, for Alzheimer's predictive modeling, it is imperative for approaches to learn unsupervised (i.e. not training on unreliable clinical labels). Several additional properties are highly desirable in medical AI in general: interpretability, uncertainty reasoning, longitudinal modeling, accounting for inconsistent sampling and missing observations in clinical data.
These qualities tend to be drawbacks of neural network approaches while being advantages of more symbolic methods~\cite{Ghassemi2020ARO, tiglic2020InterpretabilityOM}. 



\section{Problem Formulation}
\label{sec:problem}

We first define the problem of predictive modeling neurodegeneration, which is different and more clinically useful than standard survival modeling or classifying neurodegenerative disease (as discussed above).

Given previous biomarker measurements from individual patients, we aim to predict their cognitive capabilities in the future. This task avoids the unreliability of Alzheimer's clinical labels, while also being more useful in practice, as this prediction capability is valuable towards personalized prognosis and clinical trials. We aim to predict the key cognitive decline metrics for Alzheimer's progression (MMSE, ADAS-Cog13, and CDRSB) 
corresponding to length-$\tau$ horizons into the future.

Let there be $N$ subjects in a study, from which there are $K$ observation (response) variables measured at different follow-up times. The $K$ observation variables can be a mix of binary, ordinal, or continuous measurements/outcomes. We denote the measured biomarker $k$ for individual $i$ at time $j$ as $x_{ijk}$, where $i=1...N$, $k=1...K$, $j=1...q_{ik}$. 
Note $q$ is used to count the number of measurements of $k$ for individual $i$, as we must allow for the sampling to be varied and sparse.
The outcome scores for single patient visits are stored as
$\textbf{y}_{i} = \{ \textrm{MMSE} \in (0,30), \textrm{ADAS-Cog13} \in (0,85), \textrm{CDRSB} \in (0,18) \} $, and $\textbf{Y}=\{\textbf{y}_{i}\}_{i=1}^{N}$.
Furthermore, each patient is represented by data pairs: 
$\{ \textbf{x}_i, \textbf{y}_i \}$, where $\textbf{x}_i = \{x_1, ..., x_t \}$
contains the input features (i.e. biomarker values) up to visit $t$ of $T$.



\section{Proposed Method}
\label{sec:method}

Our approach flexibly combines deep neural networks, Gaussian processes, and simulation-based inference as \textit{probabilistic programmed neurodegenerative disease models}.

\subsection{Monotonic Gaussian Processes}
\label{sec:methGP}

A GP is a stochastic process which is fully specified by its mean function and covariance function such that any finite set of random variables have a joint Gaussian distribution \cite{GPML}. GPs provide a robust method for modeling non-linear functions in a Bayesian nonparametric framework; ordinarily one considers a GP prior over the function and combines it with a suitable likelihood to derive a posterior estimate for the function given data. GPs are flexible in that, unlike parametric counterparts, they adapt to the complexity of the data. Even more, GPs provide a principled way to reason about uncertainties \cite{Ghahramani2015ProbabilisticML}.

A GP defines a distribution over functions $f : \mathbb{R}^{d} \to \mathbb{R}$ from inputs to target values: 

\[
f(\mathbf{x}) \sim \mathcal{G} \mathcal{P}\left(\mu(\mathbf{x}), k_{\phi}\left(\mathbf{x}_{i}, \mathbf{x}_{j}\right)\right)
\]
\noindent
with mean function $\mu(\mathbf{x})$ and covariance kernel function $k_{\phi}\left(\mathbf{x}_{i}, \mathbf{x}_{j}\right)$ parameterized by $\phi$. Any collection of function values is jointly Gaussian,

\[
f(X) = \left[f\left(\mathbf{x}_{1}\right), \ldots, f\left(\mathbf{x}_{n}\right)\right]^{T}
\sim \mathcal{N}\left(\boldsymbol{\mu}, K_{X, X}\right)
\]
\noindent
with mean vector and covariance matrix defined by the GP, s.t. $\boldsymbol{\mu}_{i} = \mu(\mathbf{x}_{i})$ and $(K_{X,X})_{ij} = k_{\phi}\left(\mathbf{x}_{i}, \mathbf{x}_{j}\right)$. In practice, we often assume that observations include i.i.d. Gaussian noise: $y(x) = f (x) + \epsilon(x)$, where $\epsilon \sim N(0, {\phi}_{n}^{2})$, and the covariance function becomes

\[
\textrm{Cov}\left(y\left(\mathbf{x}_{i}\right), y\left(\mathbf{x}_{j}\right)\right)=k\left(\mathbf{x}_{i}, \mathbf{x}_{j}\right)+\phi_{n}^{2} \delta_{i j}
\]
\noindent
where $\delta_{i j}=\mathbb{I}[i=j]$. One can then compute a Gaussian posterior distribution in closed form by conditioning on the observed data $(X_L, \textbf{y}_L)$ and make predictions at unlabeled points $X_U$.

Sparse GP methods have significantly improved the scalability of GP inference, specifically the method of \textit{inducing points} \cite{Titsias2009VariationalLO, Hensman2015ScalableVG}. This approach allows us to make a variational approximation to the posterior process that is Gaussian and only uses $M << D$ inducing features (where $D$ is the full number of data points).
This decreases the time and space complexity from $O(d^3)$ and $O(d^2)$, respectively, to $O(dm^2)$ and $O(dm)$.

To prior the Gaussian process biomarker progressions to better model increasing severity of neurodegeneration, we implement montonicity constraints. Specifically we use the approach of \citet{Riihimki2010GaussianPW}, to include derivatives information at a number of GP input locations and force the derivative process to be positive at these locations.
In future work we would like to improve on this with monotonic GP flows \cite{Ustyuzhaninov2020MonotonicGP}.


\begin{figure*}[ht]
\centering
\includegraphics[width=1.5\columnwidth]{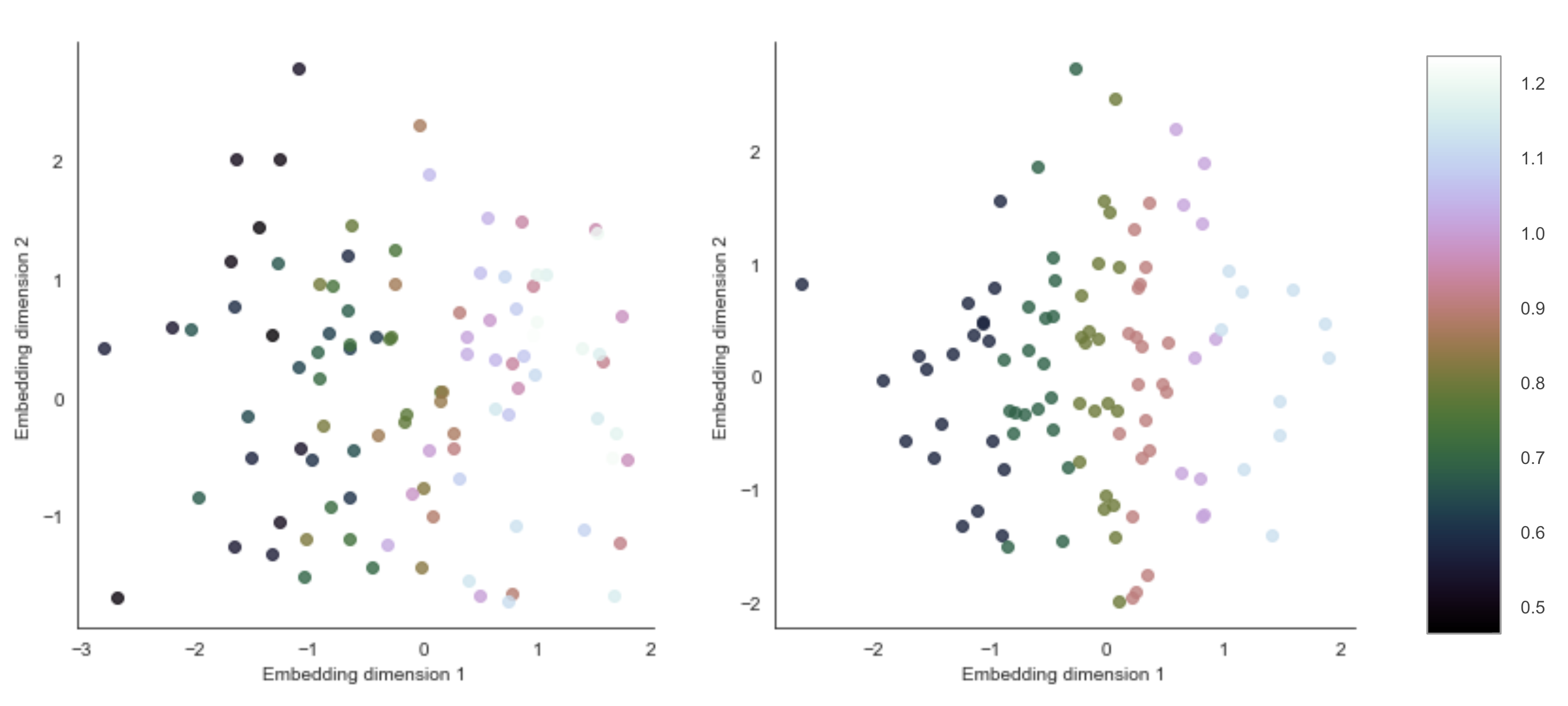}
\caption{Two-dimensional embeddings learned by a DKL model (left) and our probabilistic programmed version (right); the colorbar shows the magnitude of the normalized outputs. In the left panel, DKL learns a poor embedding, where colors representing different output magnitudes are intermingled. In the right panel, PP-DKL is able to learn a better representation of the dataset; kernel warping creates a cohesive deep kernel, yielding inducing points that lie in the original data space rather than the neural network feature space.}
\label{fig_dkl}
\end{figure*}

\subsection{Probabilistic Programmed Deep Kernel Learning}
\label{sec:PPDKL}

Deep kernel learning (DKL) combines the structural properties of deep architectures with the non-parametric flexibility of kernel methods \cite{Wilson2016DeepKL}. Specifically combining deep neural networks with Gaussian process yields several useful characteristics:
\begin{itemize}
    \item The information capacity of the model grows with the amount of available data (as a non-parametric method), but its complexity is automatically calibrated through the marginal likelihood of the GP.
    \item The non-parametric GP layer provides flexibility and automatic calibration, and typically high performance (in the face of noise). This helps reduce the need for extensive hand tuning from the user.
    \item Inference and learning procedures for O(n) training and O(1) testing time.
\end{itemize}

Given input data $\textbf{x} \in X$, a neural network parameterized by $w$ is used to extract features $h_{w}(\textbf{x}) \in \mathbb{R}$. The outputs are modeled as

\[
f(\mathbf{x}) \sim \mathcal{G} \mathcal{P}\left(\mu\left(h_{w}(\mathbf{x})\right), k_{\phi}\left(h_{w}\left(\mathbf{x}_{i}\right), h_{w}\left(\mathbf{x}_{j}\right)\right)\right)
\]
\noindent
for some mean function $\mu(\cdot)$ and covariance kernel function $k_{\phi}\left(\cdot,\cdot \right)$ with parameters $\phi$. 
Parameters $\theta = (w, \phi)$ of the deep kernel are learned jointly by minimizing the negative log-likelihood of the labeled
data: $-\log p\left(\mathbf{y}_{L} \mid X_{L}, \theta\right)$.
For Gaussian distributions, the marginal likelihood is a closed-form, differentiable expression, allowing DKL models to be trained via backpropagation.

A standard approach to DKL is to apply GPs to the feature-level outputs of a NN, one GP per feature. This representation yields GP inducing points in feature-space $H$.
\citet{Ma2019LearningIR} show kernel ``warping" to induce feature representations that respect invariances that reach beyond transformation. In this context, we would want to ``warp" the GP with the neural network feature extractor to produce a deep kernel $k_{warped}$ that yields inducing points in the space of the original data $X$:

\[
k_{warped}(x, h) = q(k(f(x), f(h)))
\]

This can be readily implemented in probabilistic programming by defining the combined deep kernel within a probabilistic program, which can then be simply used by the MCMC black-box inference engine as usual. Otherwise one must implement complex, custom inference methods for the warped deep kernel model. We refer to this approach as \textit{probabilistic-programmed deep kernel learning (PP-DKL)}.

To gain some intuition about how the warping helps the model learn a better representation, we visualize in Fig. \ref{fig_dkl} the neural network embeddings learned by the DKL and PP-DKL models on the Alzheimer's dataset (ADNI, described in the Experiments section). The left plot shows DKL learns a rather poor embedding, where different colors representing different output magnitudes are intermingled. Yet the warped deep kernel resulting from PP-DKL yields embeddings that better represent the dataset, as shown in the right plot.

\begin{table*}[!t]
\caption{Comparison of models on neurodegeneration prediction task. See text for details.}
\label{tab_results}
\begin{center}
\begin{tabular}{c | c| c| c| c| c| c}
    \hline
    \rowcolor{lightgray}
    Models & \multicolumn{3}{| c |}{MAE}  & \multicolumn{3}{ c |}{ICC}\\
    \hline
     & {MMSE \small{(0-30)}} & {A-Cog13 \small{(0-85)}} & {CDRSB \small{(0-18)}} & {MMSE \small{(0-30)}} & {A-Cog13 \small{(0-85)}} & {CDRSB \small{(0-18)}}\\
    \hline  

    \multicolumn{1}{l|}{\textbf{JM}} & $5.76\pm0.91$ & $7.20\pm1.01$ & $1.73\pm0.08$ & $0.41\pm0.20$ & $0.62\pm0.06$ & $0.74\pm$0.08\\
    \multicolumn{1}{l|}{\textbf{pGP}} & $2.18\pm0.42$ & $4.99\pm0.92$ & $0.91\pm0.11$ & $0.71\pm0.10$ & $0.79\pm0.08$ & $0.86\pm0.09$\\
    \multicolumn{1}{l|}{\textbf{PP-DKL}} & $\mathbf{1.36\pm0.27}$ & $\mathbf{3.99\pm0.09}$ & $\mathbf{0.60\pm0.06}$ & $\mathbf{0.85\pm0.11}$ & $\mathbf{0.88\pm0.06}$ & $\mathbf{0.92\pm0.04}$\\
    \multicolumn{1}{l|}{\textbf{PP-DKL'}} & $1.46\pm0.36$ & $4.04\pm0.68$ & $0.64\pm0.07$ & $0.82\pm0.13$ & $0.83\pm0.07$ & $0.91\pm0.05$\\
    \multicolumn{1}{l|}{\textbf{DKL}} & $1.82\pm0.26$ & $4.52\pm0.58$ & $0.77\pm0.05$ & $0.76\pm0.06$ & $0.79\pm0.08$ & $0.88\pm0.08$\\
    \multicolumn{1}{l|}{\textbf{Match-Net}} & $1.67\pm0.20$ & $4.71\pm0.51$ & $\mathbf{0.60\pm0.05}$ & $0.76\pm0.07$ & $0.80\pm0.10$ & $0.89\pm0.10$\\
    \multicolumn{1}{l|}{\textbf{RNN}} & $4.11\pm0.92$ & $6.96\pm0.98$ & $0.91\pm0.11$ & $0.50\pm0.19$ & $0.69\pm0.10$ & $0.75\pm1.0$\\
    \bottomrule
\end{tabular}
\end{center}
\end{table*}

\subsection{Neurodegeneration Programs}

A probabilistic model in the form of probabilistic program code describes the data-generating process from unknown latent variables values to observed data. 
We posit the underlying disease state is represented by a latent stochastic process that manifests as a series of observable, longitudinal biomarkers.
We encode additional domain knowledge by using monotonic GPs 
to model how each biomarker deteriorates over time, in agreement with main Alzheimer's progression hypotheses \cite{Jedynak2015ACM}.
The model is trained on population level data to capture general statistics for biomarker progressions.
Individual instances of a ``neurodegeneration program'' are fed a patient's observed data in the form of biomarker values, and predicts their individual biomarker progressions.

A neurodegeneration program can be used as a simulator of a patient's latent disease process. That is, probabilistic programs are likened to simulators in the sense that in simulation-based inference, the simulator itself defines the statistical model~\cite{Cranmer2020TheFO}.
The neurodegeneration program defines a statistical model with random samplings, from which we can generate probabilistic trajectories for individual biomarkers.
This is one means by which PP-DKL is interpretable: outputs can be inspected as individual and aggregate biomarker predictions over time -- i.e. trajectories with confidence intervals (from uncertainty propagation) -- rather than point predictions or a single disease risk score.
PP-DKL is also interpretable in the sense that a model is encoded as a program, so we can use the program structure and analysis to understand the properties of a model before inference~\cite{Pfeffer2018StructuredFI}.


The GP mean and variance is propagated throughout program execution, so quantifying the specific uncertainties with PP-DKL is trivial; we can sample model programs for \textit{aleatoric} (system stochasticity such as observation and process noise) and \textit{epistemic} (subjective uncertainty due to limited data) uncertainty measures. 

We implement PP-DKL in the ``Pyro'' probabilistic programming language (which is embedded in Python and leverages PyTorch) \cite{Bingham2019PyroDU}. 


\section{Experiments}
\label{sec:exp}

We experiment with data from the Alzheimer’s Disease Neuroimaging Initiative (ADNI), a multicohort longitudinal study in which volunteers diagnosed as cognitively healthy or with various degrees of cognitive impairment have been evaluated since 2005 (\url{adni.loni.usc.edu}).
ADNI subjects are evaluated via neuroimaging (PET), cerebrospinal fluid (CSF), and other biomarkers, as well as clinical and neuropsychological assessments.
As discussed earlier, the diagnostic labels in ADNI are ill-defined and thus an unsuitable task (here and in clinical practice).
We thus emphasize the task of cognitive scoring predictions corresponding to length-$\tau$ horizons into the future, rather than diagnostic classification.
We downloaded the standard dataset processed for the TADPOLE Challenge \cite{Marinescu2018TADPOLECP}, representing 1,737 unique
patients. 

To evaluate performance we follow the specifications of \cite{Peterson2017PersonalizedGP}: we ran a 10-fold patient-independent cross-validation, and report the mean absolute error (MAE) and intra-class correlation (ICC (3,1)) metrics. The latter ranges from 0–1, and is used to measure the (absolute) agreement between the model predictions and the ground-truth for target scores. Table \ref{tab_results} reports the mean $\pm$SD of the 10-folds.
All the input features were z-normalized, and we then applied principal component analysis to reduce the effects of noise in
the data, preserving 95\% of variance.
Note the experiments in \cite{Peterson2017PersonalizedGP} select a subset of patients above a certain threshold of data points. However we do not do this, as it is important to evaluate the robustness of methods to sparseness in clinical data.


\paragraph{Baselines: neural nets to symbolic models.}
We evaluate a spectrum of neurodegenerative disease modeling approaches, from neural network models with massive parameter spaces that learn nearly any relevant function approximation, to methods that are strongly-specified with domain and mechanistic details. 
At the \textit{tabula-rasa} end of the spectrum we have deep learning approaches which aim to automatically discover good feature representations through end-to-end optimization of neural networks. We specifically evaluate Match-Net \cite{Jarrett2020DynamicPI}, the leading deep learning approach in this domain, and a baseline recurrent neural networks (RNN). At the intersection of neural networks and symbolic AI, we have deep kernel learning. Specifically we compare several variations: DKL is the standard approach from \cite{Wilson2016StochasticVD}, PP-DKL is our probabilistic programmed deep kernel approach, and PP-DKL' is without the monotonicity constraints. All DKL methods use the same feature-extractor network architecture (similar to \cite{Wilson2016DeepKL}): [$d-100-50-50-2$]. Moving along the spectrum away from neural networks, we run the auto-regressive personalized GP (pGP) of \cite{Peterson2017PersonalizedGP}, and finally a conventional statistical method, joint modeling (JM). Note we do not run the purely mechanistic approach of ordinary differential equations as these have not been shown to perform well in predictive modeling neurodegenerative diseases. 
All models are implemented in PyTorch, and code will be open-source with publication.
Further details on the dataset and models are provided in the Appendix.

\subsection{Results}
\label{sec:results}

The results are shown in Table \ref{tab_results}. Notice we sort the models top-down, progressively becoming more ``pure'' deep neural network, with our neuro-symbolic approach PP-DKL listed in the middle.
We find PP-DKL in general outperforms all other methods, across the cognitive score predictions and metrics. Relative to the other deep kernel variations -- the standard DKL and the non-monotonic PP-DKL' -- we observe that both kernel-warping via probabilistic programming and the GP monotonicity constraints on biomarker progressions improve performance as expected; interestingly, the monotonic GPs show tighter variance in general.

As expected from the results in \citet{Jarrett2020DynamicPI}, we also see Match-Net outperforming the RNN baseline on the neurodegeneration prediction tasks. Interestingly, the performance difference is exacerbated in the neurodegeneration task here, which is more challenging than the fully supervised, risk-scoring task in their prior work. An important component of Match-Net that behooves longitudinal disease modeling is parameterizing a window for which to use historical measurements; comparatively, recurrent models may consume the entire history, and at the other extreme a Cox model only utilizes the most recent measurement. This is a prime example of building specific inductive biases in neural networks as a means of encoding prior knowledge (albeit minimal) for a specific task. 
Nonetheless, Match-Net (and deep learning approaches in general), are not well-suited for the task of predicting neurodegeneration as they require vast quantities of labeled data, and suffer from miscalibration and lack of interpretable mechanisms. 
PP-DKL, on the other hand, is data-efficient due to implicit modeling structure, and well-calibrated because the Bayesian formulation provides a posterior over predictions.
To evaluate the data-efficiency on this task, we find PP-DKL performs within $5\%$ of the performances in Table \ref{tab_results} when we cut the training data by $25\%$, but comparatively Match-Net errors jump upwards of $20\%$ across the tests and metrics.

Interpretability being an important feature in medical AI (for usability and trust in practice), we suggest probabilistic programming methods to be a more promising direction than ``black-box'' deep neural networks. PPL are interpretable by definition \cite{Ghahramani2015ProbabilisticML}: model structure is made explicit (rather than abstracted away and learned purely from data), and the generative formulation lends itself to model criticisms and explanation methods, such as prior and posterior predictive checks \cite{Tran2016EdwardAL}.


Further, these results have potentially important implications for early interventions in Alzheimer's disease: The known progression of neurodegenerative processes \cite{Jedynak2015ACM} suggests the longitudinal biomarkers we used as model inputs precede cognitive decline upwards of 100 months. Thus, for a model to perform well on our task, it must be able to predict cognitive decline well into the future. This result is imperative in Alzheimer's disease where interventions must target the earliest, presymptomatic phases in order to make a difference.

\section{Conclusion}
We have presented a novel probabilistic programmed deep kernel learning (PP-DKL) approach that performs state-of-the-art in predicting cognitive decline in Alzheimer's disease. Our neuro-symbolic approach has advantageous medical-AI characteristics such as data-efficiency, model and prediction interpretability, and principled uncertainty reasoning.
At a higher level, we have compared two perspectives on AI-based longitudinal disease modeling: data-driven with neural networks, and domain-engineered with probabilistic programs.

We suggest PP-DKL will perform well in other disease areas, namely Parkinson's, CTE, and other neurodegenerative conditions, which is an important direction of our future work.
Note the preference of PP-DKL over deep learning is most likely true for heterogeneous diseases with smaller datasets.

Another important direction to pursue with PP-DKL is towards diagnostic and prognostic algorithms. 
One can utilize the ``neurodegeneration programs'' as personalized disease simulators, potentially towards counterfactual reasoning and causal inference \cite{Richens2020ImprovingTA}.
Even more, the programs provide a continuous representation of each individual's disease state, i.e. representing neurodegeneration as a sample-continuous process. Further studies can thus derive an objective, data-driven definition of Alzheimer’s disease and subtypes, allowing us to stage and diagnose individuals in a precise, pathological way.

\paragraph{Broader impact.}

Artificial intelligence and machine learning methods offer a lot of promise in medicine and healthcare. Our PP-DKL method can be a powerful tool in neurodegenerative disease diagnosis and prognostic modeling; there are over 50 million people globally with Alzheimer's disease, and it is the 6th leading cause of death in the United States. The ease of experimentation with neurodegeneration models as probabilistic programs can have significant effects in better understanding neurodegenerative processes. We've made clear that Alzheimer's diagnostics are poor labels and clinical endpoints, but we can do even better than cognitive scores used in this paper and in drug development. Ideally we can utilize our approach to study a multi-biomarker model that precisely defines disease stages; cognitive scores may be noisy, prone to bias, and present difficulties with floor/ceiling effects.
Any diagnosis method, be it human or machine, in clinical practice runs the risk of errors measured in human lives. 
It is imperative that methods such as ours continue with thorough verification and validation steps (importantly, with medical professionals and other domain experts in the loop).

\bibliography{aaai21.bib}

\clearpage
\section{Appendix}

\subsection{Modeling \& Training}

We experimented with the following kernels:

\noindent The standard squared exponential (or radial basis function (RBF)) kernel:

\[
k_{\mathrm{RBF}}\left(\mathbf{x}_{i}, \mathbf{x}_{j}\right)=\phi_{f}^{2} \exp \left(-\frac{\left\|\mathbf{x}_{i}-\mathbf{x}_{j}\right\|_{2}^{2}}{2 \phi_{l}^{2}}\right)
\]

\noindent with parameters $\phi_{f}^{2}$ and $\phi_{l}^{2}$ for signal variance and characteristic length scale, respectively.

The rational quadratic kernel, which is equivalent to adding together many RBF kernels with different length scales:

\[
k_{\mathrm{RQ}}\left(x, x^{\prime}\right)=\sigma^{2}\left(1+\frac{\left(x-x^{\prime}\right)^{2}}{2 \alpha \ell^{2}}\right)^{-\alpha}
\]

\noindent where parameter $\alpha$ determines the relative weighting of large-scale and small-scale variations.

Polynomial kernels:

\[
k_{\mathrm{poly}}\left(\mathbf{x}_{i}, \mathbf{x}_{j}\right)=\left(\phi_{f} \mathbf{x}_{i}^{T} \mathbf{x}_{j}+\phi_{l}\right)^{p}, p \in \mathbb{Z}_{+}
\]

Periodic kernels and non-smooth kernels (namely Matern) would not be suitable for our biomarker progressions.

Our reported results reflect the rational quadratic kernel, although those results were marginally better than RBF.

Note we train these parameters in the log-domain to enforce positivity constraints on the kernel parameters and positive definiteness of the covariance. The parametric neural networks are regularized with L2 weight decay to reduce overfitting, and models are implemented and trained in PyTorch using the ADAM optimizer.



\subsection{Dataset Details}

Data used in this work was obtained from the Alzheimer’s Disease Neuroimaging Initiative (ADNI), a multicohort longitudinal study in which volunteers diagnosed as cognitively healthy or with various degrees of cognitive impairment have been evaluated since 2005 (\url{adni.loni.usc.edu}).
The recent TADPOLE Challenge \cite{Marinescu2018TADPOLECP} provides a standardized dataset combining data from the multiple phases of ADNI and across the study sites.
The multi-modal dataset from the ADNI database we used is comprised of biomarkers across seven different modalities:

\begin{itemize}
    \item Three cerebrospinal fluid (CSF) measurements:  amyloid-beta, tau, and phosphorylated tau levels.
    \item Cognitive assessments: CDR Sum of Boxes (CDRSB), ADAS-Cog11, ADAS-Cog13, MMSE, RAVLT-immediate subtype, RAVLT-learning subtype, RAVLT-forgetting subtype, RAVLT-percent forgetting subtype, and FAQ. We limit our experiments to only use CDRSB, ADAS-Cog13, and MMSE, which are the most prevelant in practice.
    \item Magnetic resonance imaging (MRI) biomarkers can be used to quantify atrophy and structural brain integrity by measuring the volume of specific structures, and with diffusion tensor imaging (DTI). The volumes of interest include entorhinal, fusiform, hippocampus, intracranial, mid-temp, ventricles, and whole brain.
    \item Demographics: age, gender, ethnicity, race, years of education, and marital status. All are categorical variables but for age and education.
    \item Alipoprotein E4 variant (APOE E4) gene, known as the largest generic risk factor for Alzheimer's disease. Genotyping in ADNI determines if each patient has the APOE E4 gene present, and classifies patients into one of two genetypes (APGEN1 or APGEN2) based on their alleles.
\end{itemize}

The clinical measurements are taken at approximate $1/2$-year intervals; the average absolute deviation between original values and final timestamps amounts to approximately 4 days. Some studies fix the measurements at exactly six month timestamps, but we do not.
Also note that some studies eliminate patients with sparse data, e.g. \citet{Peterson2017PersonalizedGP} cut 1,737 patients to only 100 with rich, nearly complete data. We do not eliminate patients for sparse data, as overcoming sparsity is an important capability in real-world use.

\end{document}